\documentclass{article}
\usepackage{subfigure}
\usepackage{wrapfig}
\usepackage{amsmath}
\usepackage{xspace}

\usepackage{caption}

\newcommand{\boldres}[1]{{\textbf{\textcolor{red}{#1}}}}
\newcommand{\secondres}[1]{{\underline{\textcolor{blue}{#1}}}}
\usepackage{listings}
\usepackage[]{neurips_2024}

\usepackage{graphicx}

\usepackage[utf8]{inputenc} % allow utf-8 input
\usepackage[T1]{fontenc}    % use 8-bit T1 fonts
\usepackage{hyperref}       % hyperlinks
\usepackage{url}            % simple URL typesetting
\usepackage{booktabs}       % professional-quality tables
\usepackage{amsfonts}       % blackboard math symbols
\usepackage{nicefrac}       % compact symbols for 1/2, etc.
\usepackage{microtype}      % microtypography
\usepackage{xcolor}         % colors
\usepackage{multirow}

\title{FTMixer:  Frequency and Time Domain Representations Fusion for  Time Series Modeling }

\author{%
  Zhengnan Li\\
  Department of Computer Science\\
  Cranberry-Lemon University\\
  Pittsburgh, PA 15213 \\
  \texttt{hippo@cs.cranberry-lemon.edu}
}
\author{%
  Yunxiao Qin  \\
  Department of Computer Science\\
  Cranberry-Lemon University\\
  Pittsburgh, PA 15213 \\
  \texttt{hippo@cs.cranberry-lemon.edu}
}
\author{%
  David S.~Hippocampus\thanks{Use footnote for providing further information
    about author (webpage, alternative address)---\emph{not} for acknowledging
    funding agencies.} \\
  Department of Computer Science\\
  Cranberry-Lemon University\\
  Pittsburgh, PA 15213 \\
  \texttt{hippo@cs.cranberry-lemon.edu}
}
\author{%
  David S.~Hippocampus\thanks{Use footnote for providing further information
    about author (webpage, alternative address)---\emph{not} for acknowledging
    funding agencies.} \\
  Department of Computer Science\\
  Cranberry-Lemon University\\
  Pittsburgh, PA 15213 \\
  \texttt{hippo@cs.cranberry-lemon.edu}
}

\begin{document}

\maketitle

\begin{abstract}
Time series data can be represented in both the time and frequency domains, with the time domain emphasizing local dependencies and the frequency domain highlighting global dependencies.
To harness the strengths of both domains in capturing local and global dependencies, we propose the Frequency and Time Domain Mixer (FTMixer). 
To exploit the global characteristics of the frequency domain, we introduce the Frequency Channel Convolution (FCC) module, designed to capture global inter-series dependencies. 
Inspired by the windowing concept in frequency domain transformations, we  present the Windowing Frequency Convolution (WFC) module to capture local dependencies. The WFC module first applies frequency transformation within each window, followed by convolution across windows. 
Furthermore, to better capture these local dependencies, we employ  channel-independent scheme to capture local dependencies in  time domain.
Notably, FTMixer employs the Discrete Cosine Transformation (DCT) with real numbers instead of the complex-number-based Discrete Fourier Transformation (DFT), enabling direct utilization of modern deep learning operators in the frequency domain.
Extensive experimental results across seven real-world long-term time series datasets demonstrate the superiority of FTMixer, in terms of both forecasting performance and computational efficiency.  Code is available at https://anonymous.4open.science/r/FTMixer-
\end{abstract}

\section{Introduction}
Time series data, characterized by sequential order and temporal dependencies, encapsulate valuable insights into the dynamics of various systems and processes (\cite{itransformer,liu2024timer,convtimenet,tft}). Diverse datasets such as stock prices, traffic flows (\cite{wu2021autoformer}), and ECL consumption pose unique challenges and opportunities for computational analysis (\cite{ECL, foundationsurvey,lstn}) . Each demands tailored approaches to effectively capture its inherent properties (\cite{xue2024card,liu2023nonstationary,oreshkin2020nbeats}).

\begin{figure}[htbp]
\centering
\includegraphics[scale=0.35]{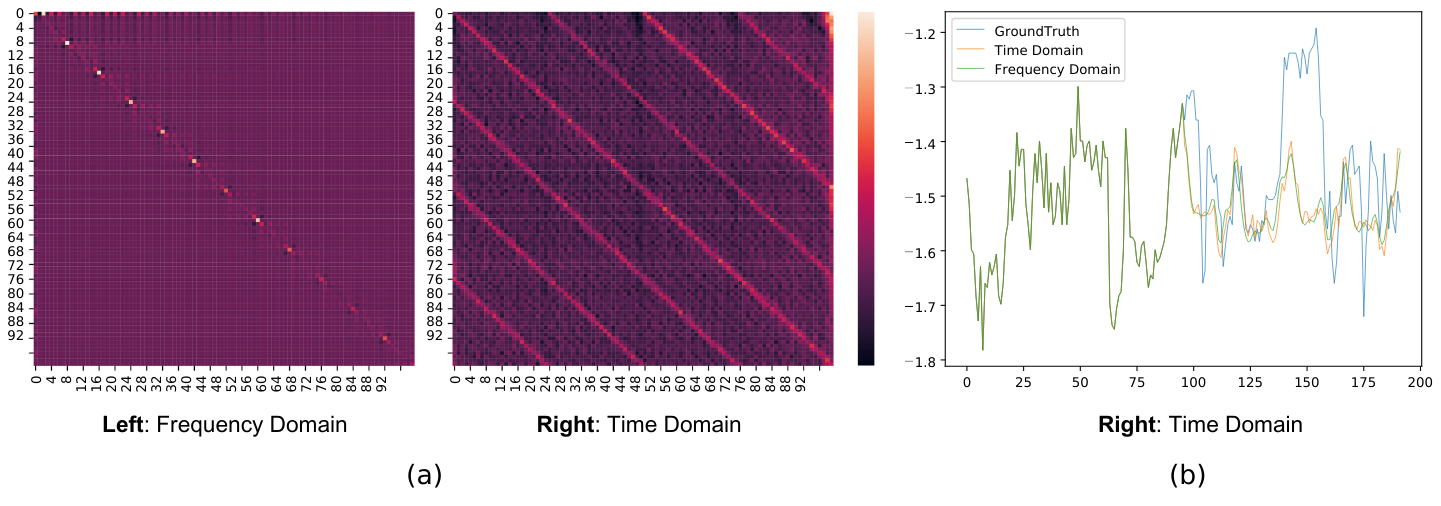}
\vspace{-5pt}
\caption{(a) Visualizations of the learned patterns of MLPs in the time domain and the frequency domain.
Learning in the frequency domain can identify clearer diagonal dependencies and key patterns than the time domain.  (b) The predictions in frequency domain and time domain. We can discover that the output in frequency domain is more smooth and focus on the low-frequency signals.  }
\vspace{-5pt}

\label{fig1}
\end{figure}

In recent years, advancements in deep learning have revolutionized time series analysis, with a primary focus on time-domain exploration. Among them, Transformer-based methods (\cite{itransformer,crossformer,fedformer,patchtst,wu2021autoformer}) and MLP-based methods (\cite{dlinear,revin,han2024softs,tide,ols}) dominate this field. Most previous methods learn time series in the time domain and achieve promising performance (\cite{qi2024pdetime}).

In the time domain, representations focus on capturing local dependencies at specific time points within the data, providing insights into how channels evolve over time and interact locally. Conversely, frequency domain representations transform the data into constituent frequencies, offering a broader perspective that enables the analysis of global dependencies and periodicities spanning the entire dataset. As depicted in Figure \ref{fig1} (a), the weights of the time domain MLP exhibit a more complex spread, with diagonal lines suggesting periodic information but overall reflecting localized, time-specific dependencies. However, the weights of the frequency domain MLP reveal sparse yet concentrated activations along the diagonal, indicating significant contributions from few frequency components and capturing broad patterns within the data. This contrast emphasizes the different strengths of each domain in capturing and representing underlying patterns within the data.

Several studies have leveraged the frequency domain to analyze time series data, enhancing feature extraction and capturing complex patterns (\cite{timesnet,fedformer,fits,fredf,frets}). For example, FEDformer (\cite{fedformer}) integrates spectral data to enhance features, while TimesNet (\cite{timesnet}) utilizes Temporal Convolutional Networks (TCNs) for feature extraction, emphasizing the significance of the Fast Fourier Transform (FFT) for periodicity extraction. Despite these advancements, the full potential of the frequency domain in time series analysis remains untapped.

However, challenges persist in developing powerful frequency-domain models. Previous frequency methods often employ Discrete Fourier Transformation (DFT) (\cite{fits,fedformer,timesnet}), which involves complex representations. Unfortunately, the widely-used techniques in deep learning, such as Batch Normalization and activation functions, on complex numbers is under-explored. Moreover, as depicted in Figure \ref{fig1} (b), the frequency domain may over-focus on global dependencies, leading to a lack of high-frequency information. Additionally, effectively combining the frequency and time domains for time series forecasting requires further exploration.

In this paper, we introduce \textbf{F}requency and \textbf{T}ime domain \textbf{Mixer} (\textbf{FTMixer}), a novel approach that efficiently incorporates both time and frequency domains to handle local and global dependencies simultaneously. 
First, to unlock the potential of the frequency domain with deep learning models, we introduce the Discrete Cosine Transformation (DCT) (\cite{dct}).
Compared to the Discrete Fourier Transform (DFT) (\cite{dct}), DCT operates exclusively on real numbers, making it more suitable for modern deep learning techniques. 
To capture inter-series global dependencies, we propose a novel Frequency Channel Convolution (FCC). The key idea of FCC is to embedding the entire sequence in the frequency domain before convolution. To better capture local dependencies, we draw inspiration from the windowing DFT (\cite{dft,fre}) and propose a novel Windowing Frequency Convolution (WFC) module. 
Specifically, WFC segments the sequence into patches with diverse window-scales and performs convolution within the patches under each scale. To aggregate the frequency domain and time domain, we propose to mix the frequency domain patches and time domain patches by Depth-Wise Separable Convolution. Finally, to better leverage the advantages of both the frequency domain and time domain, we propose to compute the loss function separately in the frequency domain and time domain.

Extensive experiments on seven datasets demonstrate that FTMixer achieves superior performance compared to state-of-the-art methods. For instance, on the ETTh1 dataset, FTMixer achieves almost a 6\% improvement over the state-of-the-art method.

\section{Related Work}
\subsection{Frequency-Aware Time Series Analysis}

Time series analysis plays a crucial role in various domains, including finance, public health, and weather forecasting (\cite{wu2021autoformer}). Recent years have witnessed  significant development in this field driven by deep learning models specifically designed for time series tasks. Among these models, three prominent architectures have garnered considerable attention: Multi-Layer Perceptrons (MLPs), Transformers, and Temporal Convolutional Networks (TCNs).

Inspired by their success in natural language processing, Transformers have been adapted for time series analysis with remarkable results (e.g., \cite{difformer}, \cite{ni2024basisformer}, \cite{atfnet}). Examples include Autoformer (\cite{wu2021autoformer}), which utilizes attention mechanisms to decompose sequences, PatchTST (\cite{patchtst}) which segments sequences inspired  by the Vision Transformer (ViT) architecture, and iTransformer (\cite{itransformer}) that embeds the entire sequence then computing attention across channel dimensions.
    
 Known for their simplicity and effectiveness, MLPs have also found application in time series analysis (e.g., \cite{revin}, \cite{liu2023koopa}, \cite{ols}, \cite{idea}, \cite{dlinear}, \cite{fits}, \cite{tide},\cite{nhits}). DLinear (\cite{dlinear}), for instance, performs trend-season decomposition and learns using two MLPs. RLinear \cite{revin} implements reversible instance norm and achieves impressive performance. Additionally, FITS (\cite{fits}) directly learns in the frequency domain, leading to surprising results.

 Temporal Convolutional Networks (TCNs) are another class of deep learning models excelling at capturing local dependencies within time series data (e.g., \cite{convtimenet}, \cite{timesnet}, \cite{micn}, \cite{moderntcn}). TimesNet (\cite{timesnet}) utilizes TCNs for feature extraction, with a particular focus on leveraging Fast Fourier Transform (FFT) for periodicity extraction. ModernTCN (\cite{moderntcn}), drawing inspiration from transformers, captures inter-series and cross-time information simultaneously. ConvTimeNet (\cite{convtimenet}) proposes a novel patch method to determine the suitable length of the patch window, enhancing the adaptability of TCNs to various time series datasets.

Despite the advancements, most existing methods capture patterns in the time domain, neglecting the potential of the frequency domain . This motivates our exploration of Frequency-Aware Time Series Analysis, which leverages insights from the frequency domain to enhance prediction accuracy.

\subsubsection{Frequency-Aware Time Series Analysis}

Several successful approaches have demonstrated the value of incorporating frequency domain information. FEDFormer \cite{fedformer} and FiLM \cite{film} integrate frequency information as additional features, enriching the model's ability to discern long-term periodic patterns while simultaneously expediting computation. TimesNet (\cite{timesnet}) adopts FFT for decomposing periodic patterns, followed by denoising via high-frequency filtering.
FreTS (\cite{frets}) forecasts time series by leveraging both inner-series and inter-series information.  
On the other hand, FITS  (\cite{fits}) achieves improved performance by training sequences directly in the frequency domain using a fully connected layer. 
However, a single linear model often proves insufficient for capturing non-linear patterns in the frequency domain. Additionally, the effectiveness of traditional deep learning techniques like activation functions and batch normalization on complex number data (used in the DFT) remains uncertain.

This work addresses these limitations by introducing the Discrete Cosine Transform (DCT) for the first time in time series analysis. 
Compared to the Discrete Fourier Transform (DFT) (\cite{dct}), DCT operates exclusively on real numbers, making it more suitable for modern deep learning techniques. 
Furthermore, DCT utilizes only amplitude to represent the frequency domain information, simplifying the computation of the loss function in the frequency domain. 
These advantages of DCT pave the way for a novel and potentially more effective approach to frequency-aware time series forecasting.

%, as it eliminates the need to represent both amplitude and phase.

% \subsection{State Space Models and Mamba}

% State Space Models (SSMs)\cite{mamba} emerge as a promising class of architectures for sequence modeling. S4 is a structured SSM where the specialized Hippo\cite{gu2020hippo} structure is imposed on the matrix $\mathbf{A}$ to capture long-range dependency. Building upon S4, Mamba~\cite{mamba} designs a selective mechanism to filter out irrelevant information and a hardware-aware algorithm for efficient implementation. Benefiting from these designs, Mamba has achieved impressive performance across modalities such as language, audio, and genomics while requiring only linear complexity on the sequence length, potentially serving as an alternative to Transformers. Due to its modeling capability and scalability, Mamba has recently shown significant progress in various communities, such as computer vision~\cite{zhu2024vision,tang2024vmrnn}, medical~\cite{ma2024u,xing2024segmamba}, graph~\cite{wang2024graph}, and recommendation systems~\cite{liu2024mamba4rec,yang2024uncovering}.

\section{Methodology}
\subsection{Prelimiary}
\subsubsection{Problem Definition}
Let $[ {X}_1, {X}_2, \cdots, {X}_T ] \in \mathbb{R}^{N \times T}$ stand for the regularly sampled multi-channel time series dataset with $N$ series and $T$ timestamps, where ${X}_t \in \mathbb{R}^N$ denotes the multi-channel values of $N$ distinct series at timestamp $t$.
We consider a time series lookback window of length-$L$ at each timestamp $t$ as the model input, namely $\mathbf{X}_t = [{X}_{t-L+1}, {X}_{t-L+2}, \cdots, {X}_{t} ] \in \mathbb{R}^{N \times L}$; also, we consider a horizon window of length-$\tau$ at timestamp $t$ as the prediction target, denoted as $\mathbf{Y}_t = [{X}_{t+1}, {X}_{t+2}, \cdots, {X}_{t+\tau} ] \in \mathbb{R}^{N \times \tau}$. 
Then the time series forecasting formulation is to use historical observations $\mathbf{X}_t$ to predict future values ${\mathbf{Y}}_t$.
%and the typical forecasting model $f_\theta$ parameterized by $\theta$ is to produce forecasting results by $\hat{\mathbf{Y}}_t = f_\theta (\mathbf{X}_t)$.
For simplicity, we shorten the model input $\mathbf{X}_t$ as $\mathbf{X} = [{X}_{1}, {X}_{2}, \cdots, {X}_{L} ] \in \mathbb{R}^{N \times L}$ and reformulate the prediction target as $\mathbf{Y} = [{X}_{L+1}, {X}_{L+2}, \cdots, {X}_{L+\tau} ] \in \mathbb{R}^{N \times \tau}$, in the rest of the paper.

\subsubsection{Discrete Cosine Transformation}

Our methodology hinges on the utilization of the Discrete Cosine Transformation (DCT) to transmute input data into the frequency domain. 
Here we elucidate the fundamental concepts of the Cosine Transformation and Fourier Transformation, forging a vital linkage between the two, before delving into the intricacies of the Discrete Cosine Transformation.

The Cosine Transform emerges as a specialized variant within the broader spectrum of the Fourier Transform. While the Fourier Transform disassembles a function into both sine and cosine constituents, the Cosine Transform singularly concentrates on the cosine components. This emphasis proves particularly advantageous in scenarios where the function exhibits symmetry, thereby facilitating a more streamlined transformational process.

The continuous Fourier Transform of a function $ f(t) $ is articulated as:
\[
F(\omega) = \int_{-\infty}^{\infty} f(t) e^{-i \omega t} \, dt,
\]
where $ F(\omega) $ symbolizes the frequency domain representation of $ f(t) $.
For an even function $ f_e(t) $, the Fourier Transform can be expressed solely in terms of cosine functions, owing to its symmetry:
\[
F(\omega) = \int_{-\infty}^{\infty} f_e(t) \cos(\omega t) \, dt.
\]

This profound connection between continuous Cosine Transformation and Fourier Transformation underscores the elegance of cosine components in symmetric functions.
Moving forward, we establish a formal relationship between the Discrete Cosine Transformation and the Discrete Fourier Transformation, enunciated through the following theorem:

\textit{\textbf{Theorem 1}: The Discrete Cosine Transform (DCT) of a sequence can be derived from the Discrete Fourier Transform (DFT) of a symmetrically extended version of the sequence.} The proof can be found in Appendix \ref{appendix:proof}.

We define the DCT for a sequence of channels as:
\begin{equation}
    \bar{x}_k = \sum_{i=1}^{L} x_i \cos\left(\frac{\pi}{L} \left(i + \frac{1}{2}\right) k\right),
    \label{eq:DCT}
\end{equation}
where $ x_i $ represents the $ i $-th element of the series $ \mathbf{X} \in \mathbb{R}^L $, and $ \mathbf{X} $ denotes the sequence in the time domain. $ L $ signifies the length of $ \mathcal{X} $, and $ \bar{x}_k $ denotes the $ k $-th frequency component in the DCT frequency domain coefficients, with $ k \in [1, 2, \ldots, L] $.

Utilizing Eq. \ref{eq:DCT}, we derive $ \bar{\mathcal{X}} = [\bar{x}_1, \bar{x}_2, \ldots, \bar{x}_L] $, representing the frequency features of $ \mathcal{X} $.

Moreover, the DCT is reversible, allowing for the transformation of frequency domain coefficients back to the time domain via the inverse Discrete Cosine Transform (iDCT), articulated as:
\begin{equation}
    x_i = \frac{1}{2} \bar{x}_1 + \sum_{k=2}^{L} \bar{x}_k \cos\left(\frac{\pi}{L} \left(k + \frac{1}{2}\right) i\right).
\end{equation}

Our proposed methodology harnesses the Discrete Cosine Transformation (DCT) to transition input data into the frequency domain. 
Renowned in signal processing, the DCT facilitates the conversion of spatial domain data into its frequency domain counterpart by emphasizing its cosine constituents. Distinguished by its operational efficiency with real numbers, the DCT mirrors the Discrete Fourier Transform (DFT) albeit with a spatial domain extension, thereby simplifying integration with deep neural networks via existing deep learning frameworks.

\subsection{Overall Architecture}
\begin{figure}
    \centering
    \includegraphics[width=0.9\textwidth]{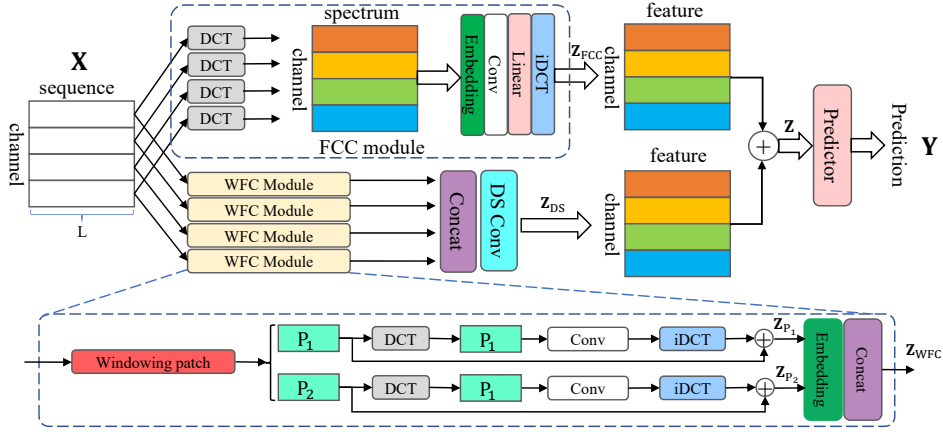}
    \caption{The framework of FTMixer. FTMixer comprises two main modules: FCC and WFC.
    An example $\mathbf{X}$ containing four channels is visualized here for easier understanding.
    The model predictor is a Linear layer.}
        \vspace{-10pt}
    \label{fig:model}
\end{figure}

To leverage both time and frequency domains to capture intricate patterns within time series data, we propose a novel method \textbf{F}requency and \textbf{T}ime domain \textbf{Mixer} (FTMixer).
% FTMixer employs a hybrid approach integrating both channel-mixing and channel-independent schemes, as illustrated in Figure \ref{fig:model}. 
% The upper pipeline represents channel-mixing, whereas the lower pipeline embodies channel-independent processing. 
% computed not only in the time domain but also the frequency domain. 
As shown in Figure \ref{fig:model}, we propose a novel Frequency Channel Convolution (FCC) module to extract inter-series information in the frequency domain.
Inspired by the concept of windowing in frequency domain transformations, we also introduce a novel module termed Windowing Frequency Convolution (WFC). 
% It firstly patchifies the input sequence with multiple window scales and then applies DCT+Conv+iDCT within each window.
% Besides, as Figure \ref{fig:model} shows, there is a skip-connection where and finally performs convolutions across windows.
Additionally, the proposed FTMixer incorporates RevIN \cite{revin} to bolster robustness. 
% The two modules enhance FTMixer's capability to capture not only local but also global dependencies within the series data. 
These components collectively form a cohesive structure that balances local and global feature extraction for time series analysis. 
In summary, the Model Structure of FTMixer can be represented as:

\begin{equation}
    \left\{
    \begin{array}{lr}
    \mathbf{Z}_{\text{FCC}} = f_{\text{FCC}}(\mathbf{X}), \\
    \mathbf{Z}_{\text{DS}} = f_{\text{DS}}(\text{Concate}(\mathcal{F}_{\text{WFC}}(\mathbf{X}) ), \\
    {\mathbf{Z} }= \mathbf{Z}_{\text{FCC}} + \mathbf{Z}_{\text{DS}}, \\
    \hat{\mathbf{Y} }= f_{\text{Pre}}(\mathbf{Z}),
    \end{array}
    \right.
    \label{eq:gradient_ensemble}
\end{equation}
where $\hat{\mathbf{Y}}$ stands for the output and $\mathbf{X}$ represents the input.
$\mathcal{F}_{\text{WFC}}$ is the mapping function that maps the WFC module to all channels in the input $\mathbf{X}$, as Figure \ref{fig:model} shows.
$f_{\text{DS}}$ and $f_{\text{Pre}}$ are the depth-wise separative convolution (DS-Conv) and the model predictor.

\subsection{Frequency Channel Convolution}

The FCC module is designed to capture inter-series dependencies in the frequency domain. However, convolution often emphasizes local dependencies, neglecting broader global dependencies due to its local invariance. 
Recognizing that the frequency domain is more conducive to revealing global dependencies compared to the time domain, we suggest embedding the entire sequence into the frequency domain prior to convolution. To achieve this, we independently apply the Discrete Cosine Transform (DCT) to each channel of the input, thereby obtaining comprehensive frequency domain representations of the sequence.

\begin{equation}
    \mathbf{X}_{f} = \text{Embedding}(\text{DCT}({\mathbf{X}}))
\end{equation}

Here, $\mathbf{X}_f$ represents the frequency domain representations of $\mathbf{X}$. Drawing inspiration from transformers, we propose to convolve along the series feature dimensions after embedding. 
%The convolution layer here is akin to a sparse Fully Connected layer, given that the embedding layer has already extracted intra-series information. 
This convolution can be formulated as:
\begin{equation}
    \vspace{-3pt}
    \mathbf{Z}_\text{FCC} = \text{iDCT}(\text{Conv1d}(\mathbf{X}_f))
\end{equation}

\subsection{Windowing Frequency Convolution}

Previous frequency-domain based models commonly focus solely on the frequency domain representation of the entire sequence, potentially leading to similar representations for two distinct sequences in the time domain. 
Drawing inspiration from the windowing concept in frequency domain transformations, we propose performing DCT within multi-scale windows to capture fine-grained information. 
Specifically, as Figure \ref{fig:model} shows, in the WFC module, we firstly segment each channel of the input sequence into patches with diverse scales and then perform DCT within each patch to obtain the local frequency domain representation. 
To capture inter-patch information, we perform convolution across these patches. 
Finally, we transform them back to the time domain and add $\mathbf{Z}_f$, which can be formulated as:
\begin{equation}
    \vspace{-3pt}
    \left\{
    \begin{array}{lr}
    \mathbf{Z}_{\mathbf{P}_j} = \text{Embedding} \Big (\text{iDCT}\big (\text{Conv}(\text{DCT}(\mathbf{P}_j))\big ) + \mathbf{P}_j \Big ), \\
    {\mathbf{Z}_\text{WFC} }= \text{Concate}(\mathbf{Z}_{\mathbf{P}_1}, \mathbf{Z}_{\mathbf{P}_2}, \ldots, \mathbf{Z}_{\mathbf{P}_n}),
    \end{array}
    \right.
\end{equation}
\vspace{6pt}
where $\mathbf{P}_j$ denotes the $j$-th patch (e.g., $\mathbf{P}_1$ or $\mathbf{P}_2$ in Figure \ref{fig:model}) in the WFC module.
$\mathbf{Z}_{\mathbf{P}_j}$ is the feature obtained for the $j$-th patch after the embedding layer.
${\mathbf{Z}_\text{WFC} }$ is the output of the WFC module for the current input channel.
Note that we separately apply the WFC module on each input channel, which can be seen in Figure \ref{fig:model}.

\subsection{Depth-Wise Separable Convolution}
After obtain ${\mathbf{Z}_\text{WFC} }$ for all input channel, we first concate them and then apply Depth-Wise Separable Convolution (DS Convolution), to further process the obtained feature.
DS Convolution decouples the learning of intra-patch dimensions from the learning of inter-patch dimensions. 
It is more computationally efficient than vanilla convolution and comprises two components: Depth-Wise Convolution (DW Conv) and Point-Wise Convolution (PW Conv). 
DW Conv aggregates inter-patch information through grouped convolution, while PW Conv operates akin to a Feed Forward Network (FFN) to extract intra-patch information. 

\subsection{Loss Function}

To thoroughly leverage the advantages of the frequency and time domains, we propose calculating the loss function specifically in each domain. 
For the time domain, we use Mean Squared Error (MSE) as the loss function, while for the frequency domain, we utilize Mean Absolute Error (MAE) loss instead of MSE.
The reason why we use L1 loss function in frequency domain is that different frequency components often exhibit vastly varying magnitudes, rendering squared loss methods unstable. 
The experimental results in Appendix \ref{sec:abl-loss} prove that both the two loss functions are helpful to improve the performance.
%Moreover, the frequency components being complex numbers pose challenges. However, since DCT is a pure real number frequency transformation, our loss function in the frequency domain
The overall loss function of our method can be expressed as:
\begin{equation}
    \vspace{-3pt}
    \left\{
    \begin{array}{lr}
    \vspace{3pt}
    \mathcal{L}_\text{time} =  \sum_{i=1}^\tau\frac{||\mathbf{Y}_i - F(\mathbf{X})_i||^2}{\tau}, \\
    \vspace{1pt}
    \mathcal{L}_\text{fre} =  \sum_{i=1}^\tau\frac{||\mathbf{Y}_i - F(\mathbf{X})_i||}{\tau}, \\
    \vspace{3pt}
    \mathcal{L}_\text{total} = \mathcal{L}_\text{time} + \mathcal{L}_\text{fre}.
    \end{array}
    \right.
\end{equation}

\vspace{-3pt}
\section{Experiment}
\vspace{-3pt}
\subsection{Experiment Settings}
\label{sec:experimental setting}
In this section, we evaluate the efficacy of FTMixer on time series  forecasting, and anomaly detection tasks. 
We show that our FTMixer can serve as a foundation model with competitive performance on these tasks. 

\begin{table}[!h]
   \vspace{-10pt}
   \caption{The Statistics of the seven datasets used in our experiments.}
   \label{t:datasets}
   \small
   \centering
      \begin{tabular}{c|ccccc}
        \toprule
         {Datasets}    & {ETTh1\&2} & {ETTm1\&2} & {Traffic} & {ECL} & {Weather}      \\ 
         \midrule
         Channels    & 7        & 7        & 862     & 321         & 21      \\
         Timesteps   & 17,420   & 69,680   & 17,544  & 26,304      & 52,696  \\
         Granularity & 1 hour   & 5 min    & 1 hour  & 1 hour      & 10 min \\
         \bottomrule
      \end{tabular}
      \vspace{-5pt}
\end{table}

\vspace{-5pt}
\paragraph{Datasets.} Our study delves into the analysis of seven widely-used real-world multi-channel time series forecasting datasets. These datasets encompass diverse domains, including ECL Transformer Temperature (ETTh1, ETTh2, ETTm1, and ETTm2) \cite{informer}, ECL, Traffic, and Weather, as utilized by Autoformer \cite{wu2021autoformer}. For fairness in comparison, we adhere to a standardized protocol \cite{itransformer}, dividing all forecasting datasets into training, validation, and test sets. Specifically, we employ a ratio of 6:2:2 for the ETT dataset and 7:1:2 for the remaining datasets. Refer to Table \ref{t:datasets} for an overview of the characteristics of these datasets.

\begin{table*}[!h]
\caption{Full forecasting results on different prediction lengths $\in \{96, 192, 336, 720\}$. Lower MSE and MAE indicate better performance. We highlight the best performance with red bold text.}
\label{tbl:forecasting_full}
\centering
\resizebox{\textwidth}{!}
{
\setlength\tabcolsep{3pt}
\begin{tabular}{c|c|cc|cc|cc|cc|cc|cc|cc|cc|cc|cc|}
\toprule

\multicolumn{2}{c}{Methods}&\multicolumn{2}{c}{FTMixer} & \multicolumn{2}{c}{ModernTCN}&\multicolumn{2}{c}{TSLANet} &\multicolumn{2}{c}{iTransformer}&\multicolumn{2}{c}{PatchTST}&\multicolumn{2}{c}{Crossformer}&\multicolumn{2}{c}{FEDformer}&\multicolumn{2}{c}{RLinear}&\multicolumn{2}{c}{Dlinear}&\multicolumn{2}{c}{TimesNet} \\

\multicolumn{2}{c}{Methods}&\multicolumn{2}{c}{------} & \multicolumn{2}{c}{ICLR 2024}&\multicolumn{2}{c}{ICML 2024} &\multicolumn{2}{c}{ICLR 2024}&\multicolumn{2}{c}{ICLR 2023}&\multicolumn{2}{c}{ICLR 2023}&\multicolumn{2}{c}{ICML 2022}&\multicolumn{2}{c}{ICLR 2022}&\multicolumn{2}{c}{AAAI 2023} &\multicolumn{2}{c}{ICLR 2023} \\
\midrule

\multicolumn{2}{c|}{Metric} & MSE  & MAE  & MSE  &MAE & MSE  & MAE  & MSE & MAE& MSE & MAE& MSE  & MAE& MSE  & MAE& MSE  & MAE& MSE  & MAE& MSE  & MAE\\
\midrule

\multirow{5}{*}{\rotatebox{90}{$ETTh1$}}
~  &  96  &  \boldres{0.356}&  \boldres{0.388} & 0.391&0.410&  0.382&  0.406 &  0.386  &  0.405  &  {0.382}&  {0.401}&  0.423  &  0.448  &  0.376  &  0.419  &  0.386  &  0.395  &  \secondres{0.375}  &  \secondres{0.399}  &  0.384  &  0.402  \\
~  &  192  &  \boldres{0.401}&  \boldres{0.410}& {0.416}&{0.423}&  0.422&  0.435 &  0.441  &  0.436  &  0.428  &  0.425  &  0.471  &  0.474  &  0.420  &  0.448  &  0.437  &  0.424  &  \secondres{0.405}  &  \secondres{0.416}  &  0.436  &  0.429  \\
~  &  336  &  \boldres{0.422}&  \boldres{0.425}& \secondres{0.437}&\secondres{0.435}&  0.443&  0.451 &  0.487  &  0.458  &  0.451  &  0.436  &  0.570  &  0.546  &  0.459  &  0.465  &  0.479  &  0.446  &  0.439  &  0.443  &  0.491  &  0.469  \\
~  &  720  &  \boldres{0.430}&  \boldres{0.454}& 0.461&0.470&  0.499&  0.498 &  0.503  &  0.491  &  \secondres{0.452}&  \secondres{0.459}&  0.653  &  0.621  &  0.506  &  0.507  &  0.481  &  0.470  &  0.472  &  0.490  &  0.521  &  0.500  \\ \cmidrule{2-22}
~  &  Avg  &  \boldres{0.402}&  \boldres{0.419}& {0.426}&\secondres{0.434}&  0.434&  0.447 &  0.454  &  0.448  &  0.428  &  0.430  &  0.529  &  0.522  &  0.440  &  0.460  &  0.446  &  0.434  &  \secondres{0.423}  &  0.437  &  0.458  &  0.450  \\

\midrule

\multirow{5}{*}{\rotatebox{90}{$ETTh2$}}
~  &  96  &  \boldres{0.275}&  \boldres{0.335}& \secondres{0.279}&0.344&  0.289&  0.351 &  0.297  &  0.349  &  0.285  &  \secondres{0.340}&  0.745  &  0.584  &  0.358  &  0.397  &  0.288  &  0.338  &  0.289  &  0.353  &  0.340  &  0.374  \\
~  &  192  &  \boldres{0.336}&  \boldres{0.375}& \secondres{0.342}&0.387&  \secondres{0.342}&  0.388 &  0.380  &  0.400  &  0.356&  \secondres{0.386}&  0.877  &  0.656  &  0.429  &  0.439  &  0.374  &  0.390  &  0.383  &  0.418  &  0.402  &  0.414  \\
~  &  336  &  \boldres{0.359}&  \boldres{0.398}& \secondres{0.363}&\secondres{0.402}&  0.371&  0.413 &  0.428  &  0.432  &  0.365&  0.405&  1.043  &  0.731  &  0.496  &  0.487  &  0.415  &  0.426  &  0.448  &  0.465  &  0.452  &  0.452  \\
~  &  720  &  \boldres{0.388}&  \boldres{0.427}& \secondres{0.390}&\secondres{0.428}&  0.417&  0.450 &  0.427  &  0.445  &  0.395  &  0.427  &  1.104  &  0.763  &  0.463  &  0.474  &  0.420  &  0.440  &  0.605  &  0.551  &  0.462  &  0.468  \\ \cmidrule{2-22}
~  &  Avg  &  \boldres{0.339}&  \boldres{0.383}& \secondres{0.344}&0.390&  0.355&  0.401
 &  0.383  &  0.407  &  0.347  &  \secondres{0.387}&  0.942  &  0.684  &  0.437  &  0.449  &  0.374  &  0.399  &  0.431  &  0.447  &  0.414  &  0.427  \\

\midrule

\multirow{5}{*}{\rotatebox{90}{$ETTm1$}}
~  &  96  &  \boldres{0.284}&  \boldres{0.334}& 0.293&0.345&  \secondres{0.286}&  \secondres{0.340} &  0.334  &  0.368  &  0.291  &  \secondres{0.340}  &  0.404  &  0.426  &  0.379  &  0.419  &  0.355  &  0.376  &  0.299  &  0.343  &  0.338  &  0.375  \\
~  &  192  &  \boldres{0.321}&  \boldres{0.361}& 0.336&0.372&  0.329&  0.372 &  0.377  &  0.391  &  \secondres{0.328}&  \secondres{0.365}&  0.450  &  0.451  &  0.426  &  0.441  &  0.391  &  0.392  &  0.335  &  \secondres{0.365}  &  0.374  &  0.387  \\
~  &  336  &  \boldres{0.355}&  \boldres{0.384}& 0.370&0.391&  \secondres{0.356}&  {0.387}&  0.426  &  0.420  &  0.365  &  0.389  &  0.532  &  0.515  &  0.445  &  0.459  &  0.424  &  0.415  &  0.369  &  \secondres{0.386}  &  0.410  &  0.411  \\
~  &  720  &  \boldres{0.415}&  \boldres{0.417}& 0.422&\secondres{0.419}&  \secondres{0.417}&  0.418 &  0.491  &  0.459  &  0.422  &  0.423  &  0.666  &  0.589  &  0.543  &  0.490  &  0.487  &  0.450  &  0.425  &  0.421  &  0.478  &  0.450  \\ \cmidrule{2-22}
~  &  Avg  &  \boldres{0.343}&  \boldres{0.373}& 0.355&0.382&  \secondres{0.347}&  \secondres{0.380}&  0.407  &  0.410  &  0.352  &  0.379  &  0.513  &  0.495  &  0.448  &  0.452  &  0.414  &  0.408  &  0.357  &  0.379  &  0.400  &  0.406  \\

\midrule

\multirow{5}{*}{\rotatebox{90}{$ETTm2$}}
~  &  96  &  \boldres{0.163}&  \boldres{0.252}& 0.168&0.257&  \secondres{0.167}&  0.262
 &  0.180  &  0.264  &  0.169  &  \secondres{0.254}&  0.287  &  0.366  &  0.203  &  0.287  &  0.182  &  0.265  &  0.167  &  0.260  &  0.187  &  0.267  \\
~  &  192  &  \boldres{0.219}&  \boldres{0.287}& \secondres{0.225}&0.297&  0.230&  0.305
 &  0.250  &  0.309  &  0.230  &  \secondres{0.294}&  0.414  &  0.492  &  0.269  &  0.328  &  0.246  &  0.304  &  0.224  &  0.303  &  0.249  &  0.309  \\
~  &  336  &  \boldres{0.269}&  \boldres{0.320}& \secondres{0.273}&\secondres{0.328}&  0.284&  0.337
 &  0.311  &  0.348  &  0.280  &  0.329  &  0.597  &  0.542  &  0.325  &  0.366  &  0.307  &  0.342  &  0.281  &  0.342  &  0.321  &  0.351  \\
~  &  720  &  \boldres{0.351}&  \boldres{0.377}& {0.370}&0.390&  \secondres{0.368}&  0.391
 &  0.412  &  0.407  &  0.378  &  \secondres{0.386}&  1.730  &  1.042  &  0.421  &  0.415  &  0.407  &  0.398  &  0.397  &  0.421  &  0.408  &  0.403  \\ \cmidrule{2-22}
~  &  Avg  &  \boldres{0.250}&  \boldres{0.309}& \secondres{0.259}&0.318&  0.262&  0.324
 &  0.288  &  0.332  &  0.264  &  \secondres{0.316}&  0.757  &  0.611  &  0.305  &  0.349  &  0.286  &  0.327  &  0.267  &  0.332  &  0.291  &  0.333  \\

\midrule

%\midrule

\multirow{5}{*}{\rotatebox{90}{$Traffic$}}
~  &  96  &  \boldres{0.362}
&  \boldres{0.238}& 0.425&0.298&  \secondres{0.372}&  \secondres{0.261}
 &  0.395  &  0.268  &  0.401  &  0.267  &  0.522  &  0.290  &  0.587  &  0.366  &  0.649  &  0.389  &  0.410  &  0.282  &  0.593  &  0.321  \\
~  &  192  &  
\boldres{0.382}
&  \boldres{0.252}& 0.435&0.302& \secondres{ 0.388}&  \secondres{0.266}
 &  0.417  &  0.276  &  0.406  &  0.268  &  0.530  &  0.293  &  0.604  &  0.373  &  0.601  &  0.366  &  0.423  &  0.287  &  0.617  &  0.336  \\
~  &  336  &  \boldres{0.389}
& \boldres{ 0.256}   & 0.446&0.306&  \secondres{0.394}& \secondres{ 0.269
}
 &  0.433  &  0.283  &  0.421  &  0.277  &  0.558  &  0.305  &  0.621  &  0.383  &  0.609  &  0.369  &  0.436  &  0.296  &  0.629  &  0.336  \\
~  &  720  &   \boldres{0.425}
&  \boldres{0.281}   & 0.452&0.311& \secondres{0.430}&  \secondres{0.289}
 &  0.467  &  0.302  &  0.452  &  0.297  &  0.589  &  0.328  &  0.626  &  0.382  &  0.647  &  0.387  &  0.466  &  0.315  &  0.640  &  0.350  \\ \cmidrule{2-22}
~  &  Avg  &  \boldres{0.390}&  \boldres{0.257}& 0.440&0.304&  \secondres{0.396}& \secondres{ 0.271}
 &  0.428  &  0.282  &  0.420  &  0.277  &  0.550  &  0.304  &  0.610  &  0.376  &  0.627  &  0.378  &  0.434  &  0.295  &  0.620  &  0.336  \\

\midrule

\multirow{5}{*}{\rotatebox{90}{$Weather$}}
~  &  96  &  
\boldres{0.143}&  \boldres{0.187}& 0.150&0.204& \secondres{ 0.148}& \secondres{0.198}
 &  0.174  &  0.214  &  0.160  &  0.204  &  0.158  &  0.230  &  0.217  &  0.296  &  0.192  &  0.232  &  0.176  &  0.237  &  0.172  &  0.220  \\
~  &  192  &  \boldres{0.188}& \boldres{ 0.232}& 0.196&0.247&  \secondres{0.194}&  \secondres{0.242}
 &  0.221  &  0.254  &  0.204  &  0.245  &  0.206  &  0.277  &  0.276  &  0.336  &  0.240  &  0.271  &  0.220  &  0.282  &  0.219  &  0.261  \\
~  &  336  &  \boldres{0.241}& \boldres{ 0.276}& 0.247&0.286&  \secondres{0.245}& \secondres{ 0.282}
 &  0.278  &  0.296  &  0.257  &  0.285  &  0.272  &  0.335  &  0.339  &  0.380  &  0.292  &  0.307  &  0.265  &  0.319  &  0.280  &  0.306  \\
~  &  720  & \boldres{ 0.318}& \boldres{ 0.332}& 0.330&0.339& \secondres{0.325}&\secondres{  0.337}
 &  0.358  &  0.349  &  0.329  &  0.338  &  0.398  &  0.418  &  0.403  &  0.428  &  0.364  &  0.353  &  0.323  &  0.362  &  0.365  &  0.359  \\ \cmidrule{2-22}
~  &  Avg  &  \boldres{0.223}& \boldres{ 0.257}& 0.231&0.269&  \secondres{0.228}& \secondres{ 0.265}
 &  0.258  &  0.278  &  0.238  &  0.268  &  0.259  &  0.315  &  0.309  &  0.360  &  0.272  &  0.291  &  0.246  &  0.300  &  0.259  &  0.287  \\
\midrule

\multirow{5}{*}{\rotatebox{90}{$ECL$}}
~  &  96  & \boldres{0.127}& \boldres{0.217}  & 0.142&0.345& \secondres{ 0.136}& \secondres{ 0.229} &  0.148  &  0.240  &  0.138  &  0.230  &  0.219  &  0.314  &  0.193  &  0.308  &  0.201  &  0.281  &  0.140  &  0.237  &  0.168  &  0.272  \\
~  &  192  & \boldres{0.145} &  \boldres{0.235}   & 0.156&0.25&  0.152&  0.244 &  0.162  &  0.253  & \secondres{0.149} & \secondres{0.243} &  0.231  &  0.322  &  0.201  &  0.315  &  0.201  &  0.283  &  0.153  &  0.249  &  0.184  &  0.289  \\
~  &  336  & \boldres{0.163} & \boldres{0.262}  & 0.174&0.269&  \secondres{0.168}& \secondres{ 0.262} &  0.178  &  0.269  &  0.169  &  0.262  &  0.246  &  0.337  &  0.214  &  0.329  &  0.215  &  0.298  &  0.169  &  0.267  &  0.198  &  0.300  \\
~  &  720  &  \boldres{0.199}& \boldres{0.285}& 0.211&0.297& { 0.205}& \secondres{ 0.293} &  0.225  &  0.317  &  0.211  &  0.299  &  0.280  &  0.363  &  0.246  &  0.355  &  0.257  &  0.331  & \secondres{0.203} &  0.301  &  0.220  &  0.320  \\ \cmidrule{2-22}
~  &  Avg  & \boldres{0.159}& \boldres{0.249
}& 0.171&0.290&  \secondres{0.165}& \secondres{ 0.257} &  0.178  &  0.270  &  0.167  &  0.259  &  0.244  &  0.334  &  0.214  &  0.327  &  0.219  &  0.298  &  0.166  &  0.264  &  0.193  &  0.295  \\
\bottomrule
\end{tabular}
}
\end{table*}

\vspace{-5pt}
\paragraph{Evaluation protocol.} Our evaluation framework, inspired by TimesNet \cite{timesnet}, hinges on two key metrics: Mean Squared Error (MSE) and Mean Absolute Error (MAE). 
To ensure a fair comparison, we adhere to a consistent evaluation protocol. 
Herein, the historical horizon's length is uniformly set at $T=96$ across all models, while prediction lengths ($F$) are selected from the set ${96, 192, 336, 720}$.

\vspace{-5pt}
\paragraph{Baseline setting.} We compare FTMixer against a variety of state-of-the-art baselines.
Transformer-based baselines include iTransformer, PatchTST, Crossformer, FEDformer, and Autoformer. 
MLP-based baselines include RLinear \cite{revin} and DLinear \cite{dlinear}. 
Besides, we also consider the Convolutional-based baseline  SCINet \cite{scinet}, the general-purpose time series models TimesNet, another frequency domain related baseline TSLANet \cite{tslanet}, and ModernTCN \cite{moderntcn}.
%, which is another frequency domain related CNN-based method.

\vspace{-5pt}
\paragraph{Inplemental Details}
Similar to the settings from \cite{onefitsall}, we set the look-back window to 336 for the ETT dataset, 512 for the Traffic,  Weather,and ECL datasets. We also incorporate the data normalization block, and reverse instance norm in the forecasting task \cite{revin}.
For the baselines, we report the best results in their original works if their settings are consistent with ours, otherwise, we re-run their official codes to obtain the performance for fair comparison. 
Notable, the official codes of some baselines contain a bug dropping the last batch in test phase.
For these baselines, we fix the bug and re-run them.
All reported results are the means under 10 random seeds.

\subsection{Experimental Results}
\textbf{Quantitative Comparison.}
Table~\ref{tbl:forecasting_full} presents the comprehensive forecasting results, with the best-performing performances highlighted in bold and the second-best highlighted in blue. Lower values of Mean Squared Error (MSE) and Mean Absolute Error (MAE) indicate better predictive performance. It is evident that FTMixer consistently demonstrates the most promising predictive performance across all datasets.
%, while FTMixer excels on almost all datasets.

To provide further insight, we select three datasets for clarification:

\textit{ETTh1 Dataset:} The dataset ETTh1 exhibits both global and complex local multi-scale dependencies. 
In this scenario, FTMixer achieves notably superior results, with an average MSE improvement of 4.9\% compared to the second-best model, PatchTST. Analysis in Table \ref{tab:ablation} indicates that both Weighted Fusion Context (WFC) and Fusion Context Vector (FCV) are effective on this dataset. Particularly, WFC plays a crucial role in capturing multi-scale local dependencies, contributing significantly to the model's performance when dealing with datasets characterized by complex local dependencies.
\textit{Weather Dataset:} This dataset is characterized by local dependencies with considerable noise. Despite these challenges, FTMixer successfully captures intricate patterns within the data, leading to significantly promising results. Patching methods, employed to effectively capture important local patterns and relationships within the time series, play a vital role in maintaining good performance. Furthermore, our WFC enhances these local patterns by effectively capturing multi-scale local dependencies.
\textit{ECL Dataset:} This dataset exhibits significant global dependencies. Even in this scenario, FTMixer outperforms other models, achieving the best performance. The robustness of FTMixer across datasets with varying characteristics underscores its efficacy in capturing both global and local dependencies effectively.
This meticulous analysis reveals the robustness and effectiveness of FTMixer across diverse datasets with different dependency structures and noise levels.

\vspace{-5pt}
\section{ Model Analysis}

\vspace{-3pt}
\subsection{Ablation Study}

\begin{wraptable}{R}{0.50\textwidth}
    \small
    \vspace{-40pt}
    \begin{tabular}{ccccc}
        \toprule
         &  ETTh1&  ECL&  Weather&  ETTm2\\
         \midrule
         w/o WFC&  0.447&  0.186&  0.285&  0.293\\
         w/o FCC&  0.427&  0.171&  0.252&  0.258\\
         Ours&  0.402&  0.159&  0.223&  0.250\\
         \bottomrule
    \end{tabular}
    \caption{The ablation experimental results about WFC and FCC.}
    \label{tab:ablation}
    \vspace{-5pt}
\end{wraptable}

In this subsection , we assess the contribution of the different components in our model, where we report the performance of the model when removing each component individually. 
The ablation study of loss function can be found in Appendix \ref{sec:abl-loss}.

\textbf{The effectiveness of WFC.} 
WFC is responsible for capturing local dependencies and improve robustness.
WFC significantly contributes to the performance on the datasets with obivious local dependencies, for instance weather and ETTh1.
As shown in Table \ref{tab:ablation}, removing WFC yields a notable decline in performance, its absence results in higher MSE values in the forecasting task of 0.447 and 0.186 for the ETTh1 and ECL datasets.
This underscores the WFC’s critical role in feature extraction  and capturing complex multi-scale local dependencies. 

\textbf{The effectiveness of FCC.}
FCC aims at capturing global dependencies and inter-series information which is  vital for performance on datasets with intricate inter-series dependencies.
As illustrated in Table \ref{tab:ablation}, removing FCC significantly harms  the performance on ECL dataset.

\vspace{-3pt}
\subsection{Computational Efficiency}
\vspace{-20pt}
\begin{wraptable}{RB}{0.62\columnwidth}
	%\begin{minipage}{\columnwidth}
    \small
    \vspace{-35pt}
    \begin{tabular}{ccccc}
        \toprule
        Cost &Benchmark & FTMixer & ModernTCN&  PatchTST \\
        \midrule
        \multirow{2}{*}{Time} &ETTh1 & \textbf{0.021s} & 0.043s & 0.027s \\
        \cline{2-5}
        & ECL & \textbf{0.023s} & 0.141s & 0.036s \\
        \midrule
        \multirow{2}{*}{Memory} & ETTh1 & \textbf{272M} & 316M & 448M \\
        \cline{2-5}
        & ECL & \textbf{304MB} & 3350MB & 828MB \\
        \bottomrule
    \end{tabular}
    \caption{The computational cost comparison between the proposed FTMixer, iTransformer, and PatchTST. The batch size for ETTh1 dataset here is $32$, while the batch size for ECL dataset here is $1$.}
    %\vspace{-3pt}
    \label{wraptable:ablation}
	%\end{minipage}
\end{wraptable}

\begin{table}[ht]
    
\end{table}

% \begin{table}
%     \centering
%     \small
%     \caption{}
    
%     \label{tab:cross-domain result}
% \end{table}
% \begin{table}
%     \begin{tabular}{|c|c|c|c|} \hline 
%          &  \textbf{FTMixer}&  iTransformer&  PatchTST\\ \hline 
%          ETTh1&  0.021s&  0.043s&  0.027s\\ \hline
%  ECL& 0.023s& 0.025s& 0.036s\\\hline
%     \end{tabular}
%     \begin{tabular}{|c|c|c|c|} \hline 
%          &  \textbf{FTMixer}&  iTransformer&  PatchTST\\ \hline 
%          ETTh1&  272M&  310M&  448M\\ \hline
%  ECL& 304M& 420M& 828M\\\hline
%     \end{tabular}
%     \caption{Caption}
%     \label{tab:time}
% \end{table}

In this subsection, we delve into the computational efficiency analysis of the proposed FTMixer. FTMixer stands out as a purely Temporal Convolutional Network (TCN)-based method, distinguished by its streamlined computational overhead. Unlike Transformer-based methodologies, which typically entail a computational complexity of $O(T^2)$ per layer, FTMixer significantly mitigates this burden to $O(KT)$, where $K$ denotes the size of the convolution kernel.
The computational complexity of the Feature Vision Convolution (FCC) component within FTMixer is expressed as $O(TMK)$, where $M$ signifies the number of channels and $K$ denotes the size of the convolution kernel. To substantiate the efficiency claims of FTMixer, experiments were conducted on the ETTh1 and ECL datasets. As delineated in Table \ref{wraptable:ablation}, FTMixer showcased superior efficiency when compared to two prominent state-of-the-art methods ModernTCN and PatchTST.
ModernTCN's efficiency diminishes as the number of channels escalates, attributable to the linear complexity of its convolution kernel concerning channel count. For instance, the ETTh1 dataset comprises 10 channels, while the ECL dataset comprises 321 channels. In stark contrast, our FTMixer component demonstrates less memory costs under identical conditions.

\vspace{-5pt}
\subsection{Visualization}
\begin{figure}
    \centering
    \vspace{-5pt}
    \includegraphics[width=\linewidth]{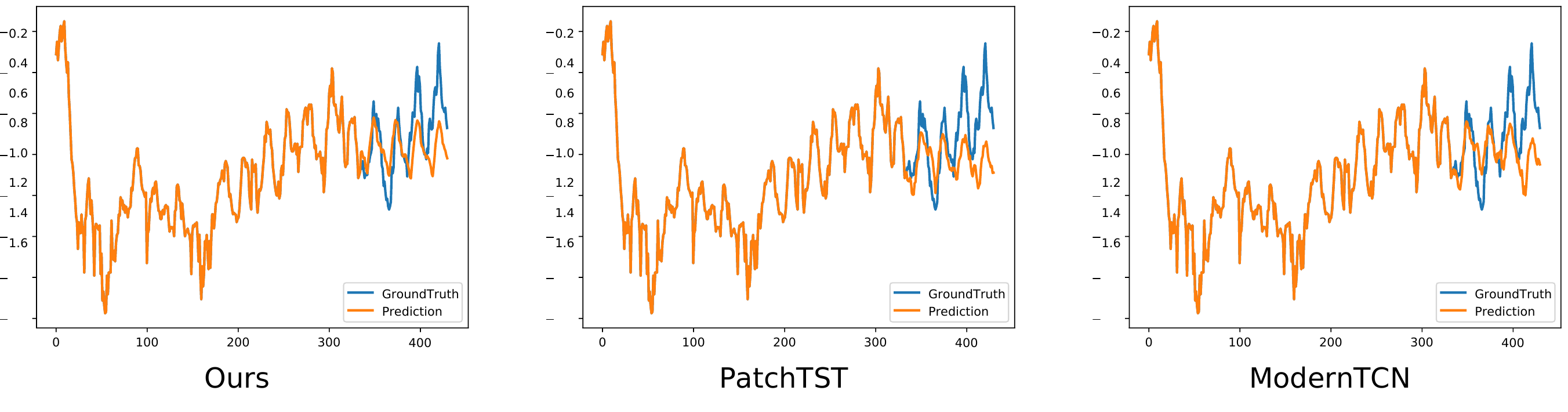}
    \vspace{-20pt}
    \caption{The visualization of the prediction of FTMixer, PatchTST, and ModernTCN on ETTh1.
    Notably, te proposed FTMixer performs the best }
    \label{fig:visualization}
    \vspace{-10pt}
\end{figure}
Here we visualize the predictions on the ETTh1 dataset. As illustrated in Figure \ref{fig:visualization}, our FTMixer model demonstrates superior prediction accuracy compared to the two baselines, PatchTST and ModernTCN. 
Unlike these time domain methods, FTMixer reveals more periodic information and effectively captures preciser global dependencies.

\begin{wraptable}{RB}{0.4\columnwidth}
	%\begin{minipage}{\columnwidth}
    \small
    \vspace{-40pt}
    \centering
    \begin{tabular}{ccccc}
    \toprule
         &  \multicolumn{2}{c}{ETTh1}&  \multicolumn{2}{c}{Weather}\\
 & MSE& MAE& MSE&MAE\\
 \midrule
         96&  0.368&  0.390&  0.165& 0.207\\
         192&  0.360&  0.389&  0.153& 0.195\\
 336& 0.357& 0.388& 0.143&0.187\\
 720& 0.355& 0.387& 0.142&0.189\\
 \bottomrule
    \end{tabular}
    \caption{The performance of the proposed FTMixer under diverse input lengths.}
    \label{tab:length}
\end{wraptable}

\vspace{-5pt}
\subsection{Varying Input Length}
In this section, we conduct experiments with varying input lengths $L \in {96, 192, 336, 720}$, while keeping the predicted length fixed at $96$. As shown in Table \ref{tab:length}, our model's performance improves with longer input lengths, highlighting the input length efficiency of our model.
A frequency component in the input may resemble a frequency component in the entire dataset. Thanks to the Wavelet Frequency Component (WFC) and Feature Vision Convolution (FCC), our model can capture the main frequency component. With longer input lengths, this frequency component becomes more representative of the entire dataset, thereby enhancing the model's length efficiency.

\vspace{-5pt}
\section{Conclusion}
\vspace{-5pt}
This paper investigates the potential of combining information from both the time domain and the frequency domain for time series forecasting tasks. We propose a novel method called FTMixer, which leverages the Discrete Cosine Transform (DCT) to integrate time-series data with its corresponding frequency domain representation. Extensive experiments demonstrate the effectiveness of FTMixer and highlight the value of frequency domain information in time series forecasting.  FTMixer achieves state-of-the-art performance while maintaining computational efficiency.
We believe these findings can stimulate greater interest in the role of the frequency domain for time series forecasting tasks.

\clearpage
%%%%%%%%%%%%%%%%%%%%%%%%%%%%%%%%%%%%%%%%%%%%%%%%%%%%%%%%%%%%
\bibliographystyle{abbrvnat}
\bibliography{references}
%%%%%%%%%%%%%%%%%%%%%%%%%%%%%%%%%%%%%%%%%%%%%%%%%%%%%%%%%%%%
\appendix
\section{Proof}
\label{appendix:proof}
\textit{\textbf{Theorem 1}: The Discrete Cosine Transform (DCT) of a sequence can be derived from the Discrete Fourier Transform (DFT) of a symmetrically extended version of the sequence.}

\textbf{Proof.}
Let $\{x_i\}_{i=1}^L$ be a real-valued sequence of length $L$. Define a new sequence $\{y_i\}_{i=1}^{2L}$ as:

\[
y_i =
\begin{cases}
x_i & \text{for } 1 \leq i \leq L, \\
x_{2L-i+1} & \text{for } L+1 \leq i \leq 2L.
\end{cases}
\]

The Discrete Fourier Transform (DFT) of a sequence of length $N$ is defined as:

\begin{equation}
Y_k = \sum_{n=0}^{N-1} y_n e^{-j2\pi kn/N}
\end{equation}

where $Y_k$ is the $k^{th}$ coefficient of the DFT, $j$ is the imaginary unit, and $e^{-j2\pi kn/N}$ is the complex exponential basis function.

The Discrete Cosine Transform (DCT) of a real-valued sequence of length $L$ (Type II DCT, for instance) is typically defined as:

\begin{equation}
C_k = \alpha_k \sum_{n=0}^{L-1} x_n \cos \left( \frac{\pi(2n+1) k}{2L} \right)
\end{equation}

where $C_k$ is the $k^{th}$ coefficient of the DCT, and $\alpha_k$ is a scaling factor that depends on the specific type of DCT.

We now show the relationship between the real part of the DFT of the extended sequence and the DCT coefficients. Since the original sequence $\{x_i\}$ is real-valued, its symmetric extension $\{y_i\}$ will also be real-valued. This property ensures that the imaginary part of the DFT of $\{y_i\}$ will be zero.

Therefore, focusing on the real part of the DFT:

\begin{align*}
\text{Re}(Y_k) &= \frac{Y_k + Y_k^*}{2}  \\
&= \frac{1}{2} \sum_{n=0}^{2L-1} y_n \left( e^{-j2\pi kn/2L} + e^{j2\pi kn/2L} \right) \\
&= \frac{1}{2} \sum_{n=0}^{L-1} x_n \left( \cos \left( \frac{\pi kn}{L} \right) + \cos \left( \frac{\pi(2L-n)k}{2L} \right) \right) \\
& \propto \sum_{n=0}^{L-1} x_n \cos \left( \frac{\pi(2n+1) k}{2L} \right) \gray{\text{Trigonometric identities}}
\end{align*}

The proportionality constant depends on the specific type of DCT (due to the scaling factor $\alpha_k$). This demonstrates that the real part of the DFT of the extended sequence is proportional to the DCT coefficients of the original sequence.

\section{The  effectiveness of Loss }
\label{sec:abl-loss}
In this section we conduct experiments to prove the effectiveness of our proposed loss function.
The input length and the output length in this experiment are $336$ and $96$.
As shown in Table \ref{tab:abl-loss},  removing the $\mathcal{L}_\text{fre}$ Loss would cause a significant performance denigration to our model.

\begin{table}[!h]
   \vspace{-10pt}
   \caption{The ablation experiments about loss function.}
   \label{tab:abl-loss}
   \small
   \centering
      \begin{tabular}{ccc}
            \toprule
            &  ETTh1& Weather\\
            \midrule
            w/o Fre Loss&  0.364& 0.146\\
            w/o Time Loss& 0.361&0.144\\
            Ours&  0.357& 0.143\\
            \bottomrule
        \end{tabular}
      %\vspace{-6pt}
\end{table}

\section{Visualization of results}
\begin{figure}
    \centering
    \includegraphics[width=1\linewidth]{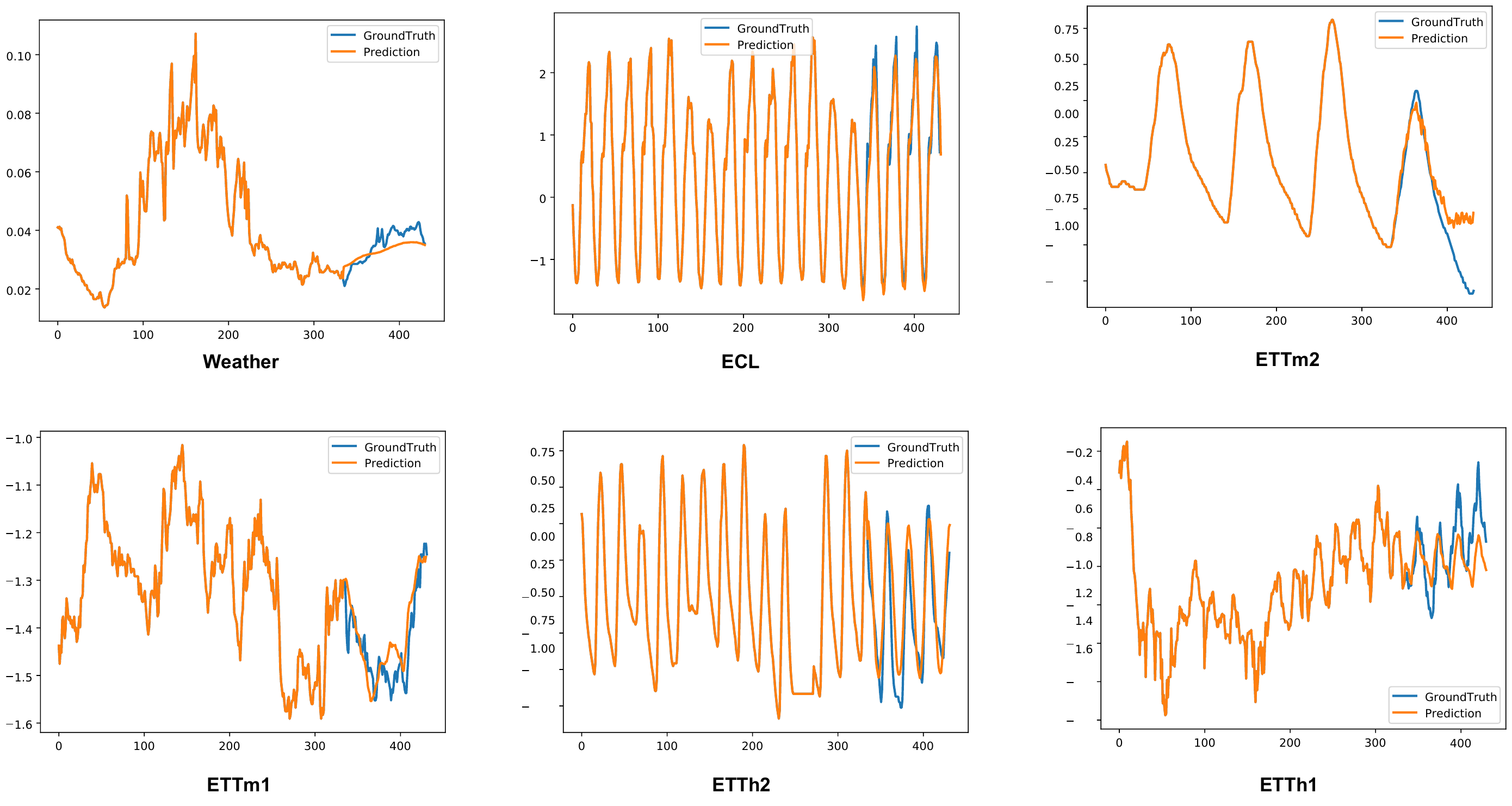}
    \caption{Enter Caption}
    \label{fig:enter-label}
\end{figure}

Figure \ref{fig:enter-label} visualizes the output of FTMixer on six real benchmarks.
FTMixer demonstrates exceptional accuracy, producing predictions highly consistent with ground truth.

\newpage
\end{document}